\documentclass[11pt]{article}

\usepackage[utf8]{inputenc}
\usepackage[T1]{fontenc}
\usepackage{lmodern}
\usepackage[margin=1in]{geometry}
\usepackage{amsmath,amssymb}
\usepackage{graphicx}
\usepackage{booktabs}
\usepackage{microtype}
\usepackage{authblk}
\usepackage{xcolor}
\usepackage[colorlinks=true,allcolors=blue!60!black]{hyperref}

\graphicspath{{figures/}{../eval/}}

\newcommand{\coderepo}{https://github.com/turangenesis/headroom}

\title{\textbf{Oversight Has a Capacity:\\ Calibrating Agent Guards to a Subjective, Fatiguing Human}}

\author{Emre Turan\thanks{Correspondence: \texttt{turangenesis@gmail.com}. Code and data: \url{\coderepo}.}}
\affil{Independent Researcher}

\date{\today}

\begin{document}
\maketitle

\begin{abstract}
As LLM agents begin to take real, irreversible actions (running shell commands, editing files,
deploying code), the standard safety pattern is a human-in-the-loop approval gate: risky actions
pause and wait for a person. We argue the gate is the easy part. The hard, unsolved part is the
\emph{judgment} --- deciding \textbf{which} actions to stop --- and the field evaluates that judgment
against two assumptions that are both false: that there is a ground-truth notion of ``risky,'' and
that the human reviewer is a perfect, infinitely-available oracle. We show, on a hand-labeled set of
125 adversarially-weighted agent actions, that (i) reviewers only \emph{moderately} agree on what is
risky (Fleiss' $\kappa = 0.52$), so there is no single correct label; (ii) framing the guard as
\textbf{selective classification under asymmetric cost} makes its operating limits measurable, and on
hard inputs the guard cannot safely auto-decide; and (iii) when the reviewer is modeled as
\textbf{endogenous} (fatiguing as escalation load grows), realized safety becomes an
\textbf{inverted-U} in the escalation rate: \emph{more human oversight can make a system less safe},
and the safety-optimal guard escalates \textbf{below full escalation} (a middle escalation rate, not
the maximum), a setting a load-aware policy also uses to resist a \emph{flooding attack} that
rubber-stamps a malicious action past a fatigued reviewer. Framed this way, agent oversight is not
only a classification problem (which action is risky?) but a resource-allocation one: human attention
is finite, and the guard's escalation policy spends it. \textbf{We claim none of these mechanisms
as novel:} fatigue-aware learning-to-defer (FALCON~\cite{falcon}), cost-sensitive deferral under workload
constraints (DeCCaF~\cite{deccaf}),
trajectory-level guarding, and fatigue/flooding attacks on human reviewers (security-operations alert
fatigue~\cite{tariq25}) are all prior art we cite. Our contribution is an \textbf{open-source
agent-oversight system} that operationalizes and \emph{measures} these ideas together in the
LLM-agent action-gating setting, turning ``is my guard good?'' from a guess into a curve. The
inverted-U and the flooding attack are modeling results that motivate a human study.
\end{abstract}

\section{Introduction}

Consider two guards for the same coding agent. Guard A escalates five actions a day to its human
reviewer; Guard B escalates five hundred. Asked which is safer, most people answer Guard B --- more
eyes on more actions, fewer dangerous calls slipping through unchecked. That answer treats the
reviewer as a fixed, reliable oracle: each escalation an independent check that can only help.

But the reviewer is not fixed. By the three-hundredth approval of a routine, benign action, a human is
fatigued and primed to keep clicking \emph{Approve} --- so a malicious action buried deep in Guard B's
stream is rubber-stamped, while Guard A's reviewer, asked only five times, reads each request
carefully. Guard B has more oversight and the worse outcome. Its escalations on routine actions were
not free: they were drawn from the same finite pool of attention the dangerous action needed.

That gap is what this paper is about. The existence of fatigue is obvious; its consequence --- that a
more-escalating oversight policy can be the \emph{less} safe one, and that the safety-optimal
escalation rate sits below ``escalate everything'' --- is not. We measure and model that tradeoff
rather than assume it. Agent oversight, framed this way, is not only a classification problem (which
action is risky?) but a resource-allocation one (how should a finite human-attention budget be spent?).

AI coding agents are moving from \emph{suggesting} code to \emph{executing} it. Once an agent can
\texttt{deploy}, \texttt{rm -rf}, push to \texttt{main}, or read a secret, a single bad instruction
has real, sometimes irreversible consequences. The de-facto safety response, echoed across agent
frameworks, is a \textbf{human-in-the-loop (HITL) approval gate}: classify each proposed action, and
pause the risky ones for a human to approve or reject.

The gate is commodity. Frameworks hand you the pause primitive (e.g.\ LangGraph's
\texttt{interrupt()}) for free. What no framework tells you is whether your approval \emph{policy} is
any good: whether it is too paranoid (so humans rubber-stamp every alert until the gate is
theater) or too lax (so something blows up). That decision is usually set by intuition.

This paper treats the decision as a measurable problem, and in doing so surfaces a deeper one. The
field implicitly assumes:
\begin{enumerate}
  \item \textbf{a ground-truth label}: that ``is this action too risky to auto-run?'' has a
  correct answer;
  \item \textbf{a perfect human}: that the reviewer the gate escalates to is always available and
  always reliable.
\end{enumerate}

Both are false. ``Risky?'' is a subjective judgment that depends on context and risk tolerance, and
human reviewers \emph{fatigue}: every escalation spends attention and nudges them toward
rubber-stamping (a well-documented ``approval fatigue'' failure mode). Crucially, the reviewer is
\textbf{endogenous}: the guard's own escalation policy degrades the very oracle it escalates to.

\paragraph{What this paper is --- and is not.} It is \textbf{not} a novel-mechanism paper.
Fatigue-aware deferral~\cite{falcon}, cost- and capacity-constrained deferral~\cite{deccaf},
trajectory-level guarding, selective classification, and
fatigue/flooding attacks on human reviewers~\cite{tariq25} are all established. It is an
\textbf{applied, measurement-driven systems} study: we bring these strands together into one
open-source agent-oversight firewall and \emph{measure} what is usually asserted. Our contributions
are therefore artifacts and measurements, not theory:
\begin{itemize}
  \item An \textbf{open-source agent firewall + measurement apparatus} that treats the guard as
  \emph{selective classification under asymmetric cost} and reports an operating-point curve,
  Neyman--Pearson point, and AURC instead of accuracy (\S\ref{sec:methods},~\S\ref{sec:calib}), with
  a live, interactive demo.
  \item A \textbf{measured noise floor} for agent-action risk: on 125 hand-labeled actions, reviewer
  agreement is only moderate (Fleiss' $\kappa = 0.52$): there is no single ground-truth safety
  label (\S\ref{sec:noise}).
  \item A \textbf{demonstration, in the LLM-agent setting, of the endogenous-reviewer
  inverted-U}~\cite{falcon}: realized safety is maximized at an escalation rate \emph{below full
  escalation}, and escalating everything is strictly worse (\S\ref{sec:invertedu}); and that the
  \emph{same} load-aware setting resists a \textbf{flooding attack}~\cite{tariq25}
  (\S\ref{sec:attack}). These are \emph{modeling} results on real scored data, not human studies.
  \item Evidence that results are \textbf{model-dependent and reproducible} (Haiku vs.\ Sonnet; AURC
  $0.374 \pm 0.002$ over seeds), and that the framework \emph{measures} both (\S\ref{sec:model}).
\end{itemize}

\noindent \textbf{Stated plainly: the contribution is a measurement instrument and the findings it
yields}: a measured noise floor where the field assumed ground truth, the operating limits of an
LLM guard framed as selective classification under asymmetric cost, and a demonstration (in the
LLM-agent action-gating setting specifically) of the endogenous-reviewer inverted-U and its
flooding-attack dual. The mechanisms are prior art; \emph{measuring} them in this setting is what we
add.

\paragraph{Scope (stated up front).} This matters where the judgment is genuinely \emph{subjective
with delayed outcomes}: autonomous agent action-gating, content-moderation borderline calls,
security-alert triage. It does \textbf{not} apply where there is objective ground truth (e.g.\
banking-fraud, eventually verifiable): there you simply measure both parties against the truth and
use the better predictor. Naming the boundary is part of the claim.

\section{Related Work}

\paragraph{Agent guardrails and trajectory-level safety.} A growing body of work guards \emph{agent
action sequences}: Trajectory Guard~\cite{traj-guard} for real-time anomaly detection over agent
trajectories, ShieldAgent~\cite{shieldagent} for verifiable safety-policy reasoning over action
trajectories, ToolSafe~\cite{toolsafe} for step-level tool-invocation guardrails, plus benchmarks
such as AgentHarm~\cite{agentharm}. \textbf{We implement per-action gating and treat trajectory-level
guarding as prior art}; our contribution is orthogonal: the \emph{oversight-calibration} layer,
which consumes whatever detection signal exists.

\paragraph{Learning to defer, complementarity, and the fatiguing expert.} A mature line studies
\emph{when} to defer to a human and how to \emph{complement} human weaknesses: learning to
defer~\cite{madras18}, complement-humans~\cite{charusaie22}, learning when to require
feedback~\cite{pugnana25}, complementary team performance~\cite{hemmer24}, appropriate
reliance~\cite{schemmer23}. Most assume a \textbf{static} expert. Two recent works move beyond pure static-expert L2D in
different ways: \textbf{FALCON}~\cite{falcon} drops the static assumption with
psychologically-grounded fatigue curves, while \textbf{DeCCaF}~\cite{deccaf} keeps static per-expert
error but adds cost-sensitive, capacity-constrained assignment. \textbf{The endogenous/fatiguing-reviewer
idea is therefore FALCON's, not ours}; the cost-and-capacity framing is DeCCaF's. We \emph{apply} both
to the LLM-agent action-gating setting and measure them.

\paragraph{Selective classification and calibration.} Risk--coverage curves and AURC come from
selective classification~\cite{geifman17}; distribution-free guarantees from conformal
prediction~\cite{angelopoulos21}; calibration is classically measured with ECE/Brier/reliability
diagrams. We use the selective-classification lens; we do \emph{not} yet claim formal calibration
(ECE). That is future rigor.

\paragraph{Reviewer fatigue as an attack surface.} That an adversary can \emph{weaponise}
alert/approval volume to exhaust reviewers and bury malicious activity is well established in security
operations (SOC alert-flooding and analyst fatigue~\cite{tariq25}) and is explicitly named as an
exploitation vector for AI agents (approval fatigue as an ``agent trap''). \textbf{The flooding
attack is therefore prior art too}; we reproduce it in the agent-oversight setting and show the
load-aware operating point defends against it (\S\ref{sec:attack}). Concurrent work reframes LLM-agent
security itself as an \emph{agent--human interaction} problem (surveying 59 papers and 21
production systems and naming \textbf{approval fatigue} and the ``cognitive burden vs.\ security''
tradeoff as first-class, under-studied concerns~\cite{wang25}), which is exactly the gap this paper
measures rather than only names. Regulatory framing of human oversight is given by EU
guidance~\cite{edps25}.

\section{Problem Formulation}\label{sec:problem}

An agent proposes actions; a guard decides, per action $a$, between \textbf{auto-allow} and
\textbf{escalate} (to a human). Let:
\begin{itemize}
  \item $s(a) \in [0,100]$ --- the guard's \textbf{risk score} (live, from rules or an LLM scorer).
  \item The \textbf{label} for $a$ is not a point but a \emph{distribution over reviewers} ---
  different reviewers disagree (\S\ref{sec:noise}). We use a gold label
  $y(a) \in \{\textsc{safe}, \textsc{approval}, \textsc{blocked}\}$ for measurement, with the noise
  floor quantifying its contestability.
  \item A threshold $\theta$: auto-allow iff $s(a) < \theta$, else escalate.
  \item An \textbf{asymmetric cost} $C[y][\text{decision}]$: auto-allowing a dangerous action (a
  \emph{miss}) is catastrophic; escalating a safe one (a \emph{false alarm}) is annoyance
  (Table~\ref{tab:cost}).
  \item A \textbf{reviewer model} $h(a, \ell)$ with reliability $r(\ell)$ that \emph{decreases} in
  cumulative escalation load $\ell$. This is the endogenous element: $\ell$ is driven by the guard's
  own escalation rate.
\end{itemize}

The objective is to minimize \textbf{expected realized cost (including human-fatigue-induced
errors)}, not classification accuracy. The endogenous element (following FALCON~\cite{falcon}) is that $r$ depends on the policy's
escalation history (a closed loop), so the optimal $\theta$ is \textbf{load-aware}, in the
cost-sensitive, capacity-constrained spirit of DeCCaF~\cite{deccaf}; we instantiate and measure this in the agent-action setting
rather than introduce it.

\begin{table}[t]
\centering
\caption{Asymmetric cost (gold $\times$ decision).}
\label{tab:cost}
\begin{tabular}{lrr}
\toprule
gold & auto-allow & escalate \\
\midrule
\textsc{safe}     & 0                 & 1 (false alarm) \\
\textsc{approval} & 5                 & 0 \\
\textsc{blocked}  & 50 (catastrophe)  & 1 \\
\bottomrule
\end{tabular}
\end{table}

\section{Methods}\label{sec:methods}

\paragraph{Dataset.} 125 hand-labeled agent actions (\texttt{eval/dataset.jsonl}), deliberately
weighted to \emph{hard} cases, including 54 ambiguous-middle (e.g.\ \texttt{npm install <pkg>}, edit
\texttt{package.json}, \texttt{git rebase}, deploy to \emph{staging}), 23 obfuscated/adversarial
(base64-encoded \texttt{rm -rf}, homoglyph \texttt{ma\'{i}n}, path traversal, secret exfiltration,
pipe-to-shell), and 16 scary-\emph{looking} cases (12 genuinely benign false-alarm traps:
\texttt{rm -rf node\_modules}, read \texttt{.env.example}; and 4 that look alarming \emph{and} warrant
approval, e.g.\ \texttt{git reset -{}-hard} of unpushed work), alongside clearer allow/block cases that anchor the easy end. Labels: \textsc{safe} 42 /
\textsc{approval} 52 / \textsc{blocked} 31. A small, curated set, reported as such, not a published
benchmark. \textbf{Gold-label provenance:} the labels are one author's judgments, used as a single
measurement reference throughout. This is in deliberate tension with our own noise floor
(\S\ref{sec:noise}): Fleiss' $\kappa = 0.52$ quantifies how contestable that reference is, so every
guard score below should be read \emph{relative to} that floor, not as agreement with an objective
truth.

\paragraph{Guard scorer.} Deterministic rules score clear cases for free; the ambiguous middle is
scored by an LLM (Haiku by default, Sonnet for comparison), at temperature 0, prompted for a 0--100
risk integer. Scores are persisted so analyses replay them without re-querying.

\paragraph{Calibration.}\label{sec:calibdef} A \emph{dangerous} action is one whose gold label is in
$\{\textsc{approval}, \textsc{blocked}\}$; the binary decision is \textbf{auto-allow}
($s(a) < \theta$) vs.\ \textbf{escalate}. Sweeping $\theta$ we report, per operating point:
\textbf{missed-danger rate} (dangerous actions auto-allowed / all dangerous), \textbf{false-alarm
rate} (safe actions escalated / all safe), \textbf{coverage} (auto-decided fraction), and
\textbf{expected cost} $=$ mean over all actions of the Table-\ref{tab:cost} cost
$C[\text{gold}][\text{decision}]$. From the sweep we extract the cost-minimizing point, the
Neyman--Pearson point (lowest false-alarm at 0\% miss), and the \textbf{AURC} $=$ area under the
risk--coverage curve, where \emph{risk} is the error rate among auto-allowed actions and
\emph{coverage} is the auto-allowed fraction (lower is better). (\texttt{eval/calibrate.py}.)

\paragraph{Noise floor.} Three reviewer \emph{personas} (cautious / pragmatic / strict-compliance)
label the set; we compute pairwise Cohen's $\kappa$ and overall Fleiss' $\kappa$. \textbf{These are
LLM personas, a proxy for human annotators, reported as such.} (\texttt{eval/noise\_floor.py}.)

\paragraph{Endogenous-reviewer simulation.} We model reviewer reliability
\[
  r(\ell) = \max\!\big(r_{\min},\; 1 - \text{slope}\cdot\max(0,\, \ell - C)\big)
\]
with capacity $C$, $\text{slope} = 0.02$, $r_{\min} = 0.2$: the reviewer is reliable up to $C$
reviews, then degrades. For each $\theta$, auto-allowed dangerous actions are guard-misses; escalated
dangerous actions are missed with probability $1 - r(\ell)$ at their load position. We sweep $\theta$
(hence the escalation rate) and vary $C$. \textbf{This models a documented phenomenon; it is not a
human study.} (\texttt{eval/inverted\_u.py}.)

\section{Experiments and Results}\label{sec:results}

\subsection{The guard's judgment is measurable --- and limited on hard inputs}\label{sec:calib}

\begin{figure}[t]
\centering
\includegraphics[width=\linewidth]{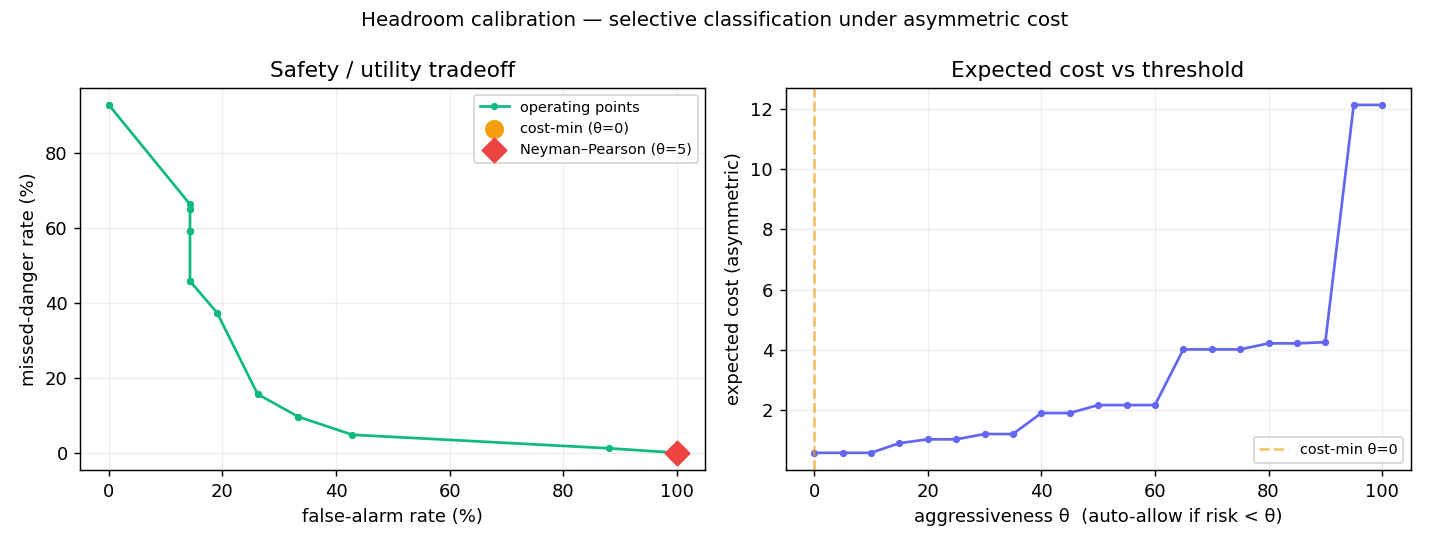}
\caption{Safety/utility tradeoff (left) and expected cost vs.\ threshold (right) for the LLM-scored
guard on the 125-action set, with the cost-minimizing and Neyman--Pearson points marked.}
\label{fig:calib}
\end{figure}

On the 125-action set the guard's safety/utility tradeoff is a real curve, not a binary. Under the
asymmetric cost (Table~\ref{tab:cost}), the \textbf{cost-minimizing policy collapses to ``escalate
almost everything''}: reaching 0\% missed-danger requires a $\sim$100\% false-alarm rate, and the
area under the risk--coverage curve (AURC; lower is better) is \textbf{0.376} for the run shown,
and \textbf{$0.374 \pm 0.002$} across three temperature-0 seeds (\S\ref{sec:model}), the figure we
quote as canonical throughout. The reading is not ``the guard is bad''; it is that \emph{on
adversarial/ambiguous inputs this guard cannot safely auto-decide}, so it is forced to lean on the
human. That dependence is precisely what makes the reviewer's properties decisive.

\subsection{There is no single ground truth (noise floor)}\label{sec:noise}

Three persona reviewers labeling the same 125 actions reach only \textbf{Fleiss' $\kappa = 0.52$}
(moderate agreement). The three pairwise Cohen's $\kappa$ are \textbf{0.42} (cautious vs.\ pragmatic),
\textbf{0.47} (pragmatic vs.\ compliance), and \textbf{0.71} (cautious vs.\ compliance) --- i.e.\ two
of three pairs are only weak-to-moderate, and the \emph{pragmatic} reviewer labels 87 actions
\textsc{safe} versus the cautious reviewer's 45. The disagreement is concentrated on the
risk-tolerance axis, exactly the contested middle. The persona majority matches the gold label 74\%
of the time. A guard cannot be scored against one objective truth; the agreement ceiling is the
honest yardstick. \emph{(Personas are a proxy for human annotators.)}

\subsection{Oversight has a capacity (the inverted-U)}\label{sec:invertedu}

\begin{figure}[t]
\centering
\includegraphics[width=\linewidth]{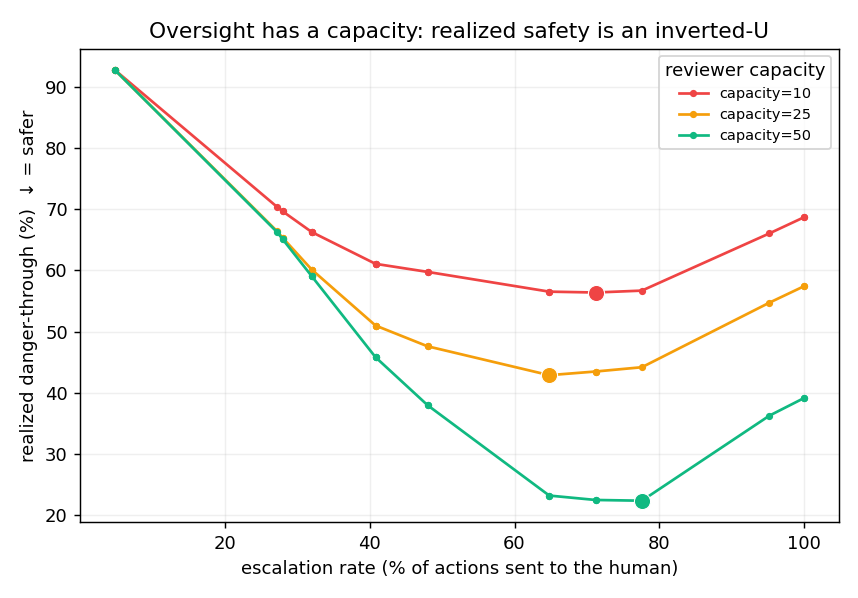}
\caption{Realized danger-through vs.\ escalation rate for three reviewer capacities; the marked minima
are the safety-optimal operating points (all below full escalation).}
\label{fig:invertedu}
\end{figure}

Modeling the reviewer as endogenous flips the usual intuition. \textbf{We state this up front: the
inverted-U below is a direct consequence of the assumed monotonically-fatiguing reviewer
(\S\ref{sec:methods}) --- a modeling result about a plausible model, not an empirical finding about
real people.} As the guard escalates \emph{more}, two
failure modes trade off: escalate too little and the guard auto-allows danger (guard-misses); escalate
too much and the reviewer overloads and rubber-stamps (fatigue-misses). Realized danger-through is
therefore \textbf{U-shaped in the escalation rate} --- and the safety-optimal escalation rate is
\textbf{below full escalation}:

\begin{table}[t]
\centering
\small
\caption{Safety-optimal escalation is below full escalation, and the optimum shifts with capacity.}
\label{tab:invertedu}
\begin{tabular}{rrrr}
\toprule
reviewer capacity $C$ & optimal esc.\ rate & danger-through @ optimum & danger-through @ full esc. \\
\midrule
10 & 64\% & 56\% & 69\% \\
25 & 64\% & 42\% & 57\% \\
50 & 72\% & 22\% & 39\% \\
\bottomrule
\end{tabular}
\end{table}

\textbf{Escalating everything is strictly worse than the optimum}, and the optimum shifts with
capacity. The absolute danger-through is high because the guard is weak on this hard set
(\S\ref{sec:calib}): the claim is the \emph{shape} (more oversight $\rightarrow$ less safety past
the optimum), not the level. This is a \emph{modeling} result; a human study fitting $r(\ell)$ (future
work) would make it empirical.

\subsection{The result is model-dependent --- and we measure it}\label{sec:model}

\begin{figure}[t]
\centering
\includegraphics[width=\linewidth]{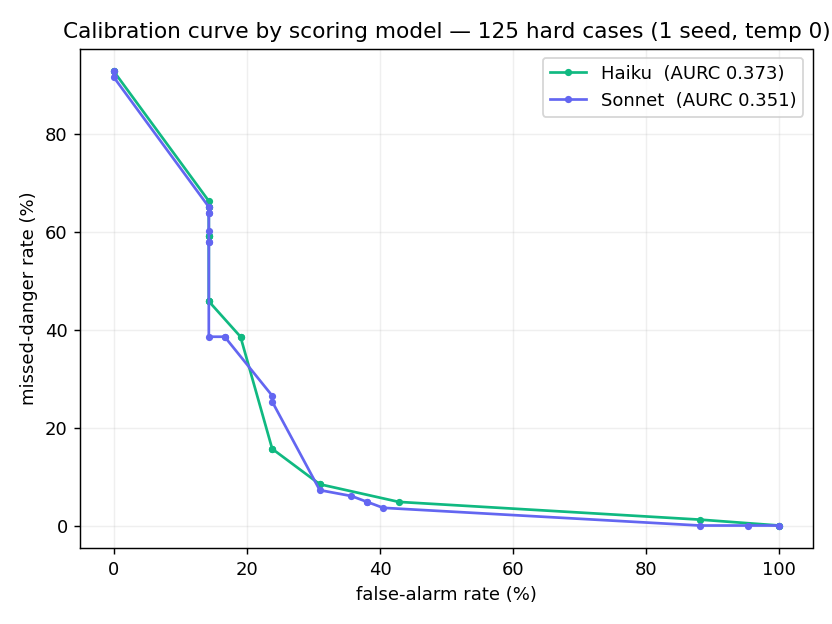}
\caption{Calibration curves for two scoring models (Haiku vs.\ Sonnet) on the 125-action set.}
\label{fig:model}
\end{figure}

Re-scoring with a stronger model shifts the curve: \textbf{Sonnet AURC 0.351 vs.\ Haiku 0.373} (each a
single seed; the Haiku value sits inside the $0.374 \pm 0.002$ band below), and Sonnet can safely
auto-allow a sliver (cost-min $\theta = 10$) where Haiku must escalate everything ($\theta = 0$). The
gain is \emph{modest}: a better model helps but does not solve the hard set, and the $0.022$ AURC
gap is small enough that it may sit within the label subjectivity the noise floor measures
(\S\ref{sec:noise}, $\kappa = 0.52$): we report the \emph{ordering}, not a precise magnitude. The point is
methodological: guard quality depends on the scoring model (and threshold, and attack mix), so the
right output is not ``guards are good/bad'' but \emph{a measurement, for a given configuration}. The
result is also \textbf{reproducible at the deployed setting}: across 3 runs at temperature 0 the
Haiku AURC is \textbf{$0.374 \pm 0.002$} (range 0.372--0.378) --- note that even at temperature 0 the
Anthropic API is not bit-exact, so we report the residual spread rather than asserting determinism;
temperature 0.7 gives the same $0.374 \pm 0.002$. The curve is stable to LLM sampling, not a single
lucky draw.

\subsection{Fatigue is an attack surface}\label{sec:attack}

\begin{figure}[t]
\centering
\includegraphics[width=\linewidth]{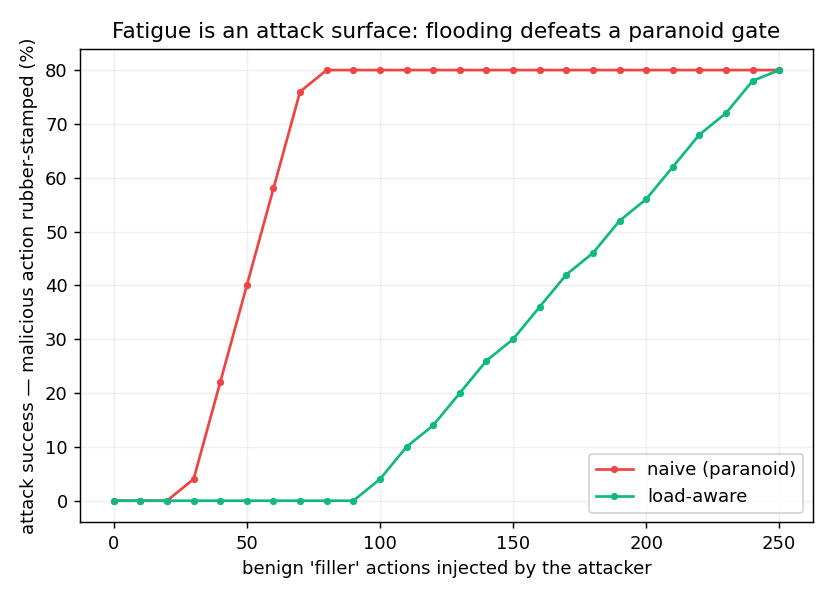}
\caption{Attack success (a buried malicious action rubber-stamped) vs.\ attacker filler volume, under
a paranoid vs.\ a load-aware escalation policy (capacity $= 25$).}
\label{fig:attack}
\end{figure}

The endogenous reviewer is not only a safety limit --- it is an \emph{attack surface}. An adversary
buries one malicious action in a stream of $k$ benign ``filler'' actions; each escalated filler adds
reviewer load, so by the time the malicious action is reviewed a fatigued reviewer rubber-stamps it.
Under a naive \textbf{paranoid} policy --- escalating 88\% of routine actions (the measured false-alarm
rate at $\theta = 10$ for the canonical Haiku guard) --- attack success climbs from 0 toward the rubber-stamp ceiling ($\sim$80\%)
as filler grows, and is already \textbf{40\% at just 50 filler actions}. Under a \textbf{load-aware}
policy --- escalating 26\% (the $\theta = 35$ false-alarm rate) --- the reviewer stays fresh and attack
success \textbf{holds at 0\% until the attacker's filler is large enough that even those few
escalations exceed capacity} ($\approx$100 actions here, $\sim$2$\times$ the paranoid policy's
headroom), after which it too bends upward (10\% at $k \approx 110$, reaching the same ceiling by
$k \approx 250$). The defense buys \textbf{headroom, not immortality} --- but it is the \emph{same
lever} as the safety result: \textbf{not escalating routine actions} both finds the inverted-U optimum
(\S\ref{sec:invertedu}) and denies the attacker the load they need. (Simulation, same fatigue model as
\S\ref{sec:invertedu}.)

\subsection{Threats to validity}\label{sec:threats}

Four sensitivities shape how far these results should be read. \textbf{(i) The cost matrix drives the
\S\ref{sec:calib} collapse.} The 50:1 \textsc{blocked}-miss-to-false-alarm ratio in Table~\ref{tab:cost}
is a chosen value; a less extreme ratio would move the cost-minimizing point away from ``escalate
everything.'' We fix it for all runs and report it explicitly rather than tuning it. \textbf{(ii) The
reviewer model is optimistic.} $r(\ell)$ starts at 1.0 (a perfectly reliable fresh reviewer) and only
decays: real reviewers err even when fresh, so the \emph{absolute} danger-through numbers are lower
bounds; the \emph{shape} (the inverted-U) is what we claim. \textbf{(iii) Agreement is weak on two of
three pairs} (\S\ref{sec:noise}), so the $\kappa = 0.52$ floor is itself an estimate over contestable
labels. \textbf{(iv) One dataset underlies every experiment.} All five results use the same 125
actions; they are five views of one measurement instrument on one small set, not five independent
confirmations.

\section{Limitations}\label{sec:limitations}

\begin{itemize}
  \item \textbf{Small, curated dataset} (125 actions), single domain (coding-agent actions); results
  are demonstrations of the \emph{instrument}, not population estimates.
  \item \textbf{Personas are a proxy} for human annotators; the $\kappa = 0.52$ floor is an estimate,
  not the true human-agreement ceiling.
  \item \textbf{The inverted-U is simulated}, not measured: fatigue is documented, but $r(\ell)$'s
  shape is assumed, not fit to people.
  \item \textbf{Sampling sensitivity is small but nonzero even at the deployed temperature 0} (AURC
  $0.374 \pm 0.002$ over 3 runs; the Anthropic API is not bit-exact at temp 0) --- quantified, not
  assumed.
  \item \textbf{Operating-point analysis, not formal calibration} (no ECE/reliability yet).
  \item \textbf{The core mechanisms are prior art} (verified): fatigue-aware
  deferral~\cite{falcon}, cost-/capacity-constrained deferral~\cite{deccaf}, and reviewer-fatigue
  attacks~\cite{tariq25} are established. This is
  an \emph{applied/measurement/systems} study, not a theoretical contribution --- positioned as such,
  not as a discovery.
\end{itemize}

\section{Future Work}\label{sec:future}

The apparatus here is a foundation, not an endpoint. Four directions extend it most directly; each is
\textbf{open direction, not result}, and each builds on what is already built.

\paragraph{7.1 From modeled fatigue to measured fatigue (the empirical core).} The inverted-U is
simulated; the obvious next artifact is a \textbf{human study} that fits $r(\ell)$ from real reviewers
and re-measures the noise floor with human annotators rather than LLM personas. Even a small pilot
would convert the headline from ``under a plausible fatigue model\ldots'' to ``under \emph{measured}
fatigue\ldots'' --- and would yield, to our knowledge, the first empirical fatigue curve for
\emph{agent-action} review specifically. It also motivates a \textbf{learned, value-of-review
deferral} policy that escalates by the expected benefit of review under the remaining attention
budget, rather than by a risk threshold alone.

\paragraph{7.2 From cooperative to enforced oversight (the no-bypass ladder).} Our MCP integration is
\emph{cooperative} --- the agent is configured to ask. A control plane is only as strong as the
actions it actually sees, so the foundational open problem is \textbf{interception completeness}: an
MCP \textbf{gateway} (the agent's only path to its tools), \textbf{host hooks} (e.g.\ Claude Code
\texttt{PreToolUse}, covering native tools too), and ultimately \textbf{capability/sandbox mediation},
where the agent runs without the real filesystem/network/deploy capability and the guard mediates the
syscall. Only the last gives \emph{true} no-bypass: enforcement is owning the chokepoint, not
requesting cooperation.

\paragraph{7.3 From per-action to trajectory and multi-agent risk.} A single action can be safe while
the \emph{sequence} is lethal (read secret $\rightarrow$ write public file $\rightarrow$ push);
trajectory-level guarding~\cite{traj-guard,shieldagent} is the detection layer this oversight
calibration would consume. Further out lies a genuinely under-explored frontier: \textbf{emergent
multi-agent risk}, where $N$ agents are each individually compliant but their \emph{joint} action
space is dangerous. Calibrating oversight across a fleet is also where fatigue compounds: one
reviewer cannot be the oracle for many agents.

\paragraph{7.4 From a static to a self-improving policy.} The guard's own \textbf{audit log is a
training signal}: which actions a reviewer \emph{always} approves vs.\ \emph{always} blocks, and which
adversarial cases slipped through. A closed loop --- adapt thresholds from the approve/reject record
and let a model \emph{propose} new rules from the misses, re-evaluated before they ship --- would make
the gate learn from every human decision and every attack it let slip.

Beyond these, standard rigor extensions apply: \textbf{conformal} abstain-or-act guarantees and formal
\textbf{calibration} (ECE), scaling to \textbf{published benchmarks} (AgentDojo, InjecAgent) for
external validity, and --- where measured reliability warrants it --- \textbf{consented, revocable}
delegation of specific decision classes to the guard (\S\ref{sec:ethics}).

\section{Ethical Considerations}\label{sec:ethics}

The framing here is \textbf{decision support for the operator}, not replacement of human authority. A
natural extension --- measured comparative reliability leading the human to \emph{delegate} certain
decision categories to the guard --- must be \textbf{consent-based, revocable, and category-scoped}:
the operator chooses, with data, and can revoke. We explicitly avoid any framing in which an agent
overrides a person's judgment without consent. The fatigue result also has a defensive reading:
because rubber-stamping is exploitable, modeling it is a step toward \emph{protecting} reviewers, not
automating them away.

\section{Conclusion}

Stopping an agent is a framework feature. Knowing \emph{when} to stop it --- and accounting for the
fact that asking depletes the human you are asking --- is the problem. A guard that escalates more is
not automatically safer: past the reviewer's capacity, the extra escalations spend the very attention
the dangerous action will need. Treating the guard as selective
classification under asymmetric cost makes its judgment measurable; measuring reviewer agreement shows
there is no single ground truth; and modeling the reviewer as endogenous shows that oversight has a
capacity, beyond which more of it makes a system less safe. None of these mechanisms is ours to claim
--- fatigue-aware deferral and reviewer-fatigue attacks are prior art --- and we say so. What we
contribute is the \textbf{system and the measurement}: an open-source agent firewall that brings these
strands together and turns ``is my guard any good?'' from a vibe into a curve. The numbers are small
and some are simulated, and we say that too.


\end{document}